\documentclass{article}
\usepackage[T1]{fontenc}
\usepackage[utf8]{inputenc}
\usepackage[]{ismir} 
\usepackage{amsmath,amsfonts,cite,url}
\usepackage{graphicx}
\usepackage{color}
\usepackage{booktabs}
\usepackage{amsmath}

\usepackage{multirow}
\usepackage{booktabs}
\usepackage{arydshln}
\usepackage{makecell}
\usepackage{float}

\newcommand{\muskern}{\kern-.15ex } 

\makeatletter
\newcommand\dynmark[1]{{\normalfont\bfseries\itshape
  \@tfor\next:=#1\do{\put@muskern\next}\/}}
\newcommand{\put@muskern}{\let\put@muskern\muskern}
\makeatother

\usepackage{acronym}
\acrodef{SMB}{Sheet Music Benchmark}
\acrodef{OMR}{Optical Music Recognition}
\acrodef{CWMN}{Common Western Modern Notation}
\acrodef{LA}{Layout Analysis}
\acrodef{SER}{Symbol Error Rate}

\usepackage{enumitem}
\setlist[itemize]{align=parleft,left=0pt,label={}}

\title{Sheet Music Benchmark:\\Standardized Optical Music Recognition Evaluation}

\multauthor
  {Juan C. Martinez-Sevilla$^1$ \hspace{1cm} Joan Cerveto-Serrano$^1$ \hspace{1cm} Noelia Luna$^1$}
  {{\bf Greg Chapman$^2$ \hspace{1cm} Craig Sapp$^3$ \hspace{1cm} David Rizo$^{1, 4}$ \hspace{1cm} Jorge Calvo-Zaragoza$^1$}\\
  $^1$ Pattern Recognition and Artificial Intelligence Group, University of Alicante, Spain\\
  $^2$ Self-employed\\
  $^3$ Center for Computer Research in Music and Acoustics, Stanford University, USA\\
  $^4$ Instituto Superior de Enseñanzas Artísticas de la Comunidad Valenciana, Spain\\
  {\tt\small \{jcmartinez.sevilla, joan.cerveto, noelia.luna, drizo, jorge.calvo\}@ua.es} \\{\tt\small gregc@mac.com, craig@ccrma.stanford.edu}
  }

\def\authorname{J. C. Martinez-Sevilla, J. Cerveto-Serrano, N. Luna, G. Chapman, C. Sapp, D. Rizo, and J. Calvo-Zaragoza}

\usepackage[bookmarks=false,pdfauthor={\authorname},pdfsubject={\pdfsubject},hidelinks]{hyperref}

\begin{document}

\maketitle

\begin{abstract}
In this work, we introduce the Sheet Music Benchmark (SMB), a dataset of six hundred and eighty-five pages specifically designed to benchmark Optical Music Recognition (OMR) research. SMB encompasses a diverse array of musical textures, including monophony, pianoform, quartet, and others, all encoded in Common Western Modern Notation using the Humdrum \texttt{**kern} format. Alongside SMB, we introduce the OMR Normalized Edit Distance (OMR-NED), a new metric tailored explicitly for evaluating OMR performance. OMR-NED builds upon the widely-used Symbol Error Rate (SER), offering a fine-grained and detailed error analysis that covers individual musical elements such as note heads, beams, pitches, accidentals, and other critical notation features. The resulting numeric score provided by OMR-NED facilitates clear comparisons, enabling researchers and end-users alike to identify optimal OMR approaches. Our work thus addresses a long-standing gap in OMR evaluation, and we support our contributions with baseline experiments using standardized SMB dataset splits for training and assessing state-of-the-art methods.
\end{abstract}

\section{Introduction} \label{sec:introduction}

\ac{OMR} is a long-standing challenge within the field of Music Information Retrieval (MIR). It focuses on automatically extracting musical information from scanned images, manuscripts, or printed documents, converting this information into structured digital representations~\cite{Calvo-ZaragozaH20understanding}, such as Humdrum \texttt{**kern}~\cite{Huron1997}, MEI~\cite{HankinsonRF11}, or MusicXML~\cite{good2001musicxml}. These machine-readable formats facilitate large-scale musical information retrieval, enable advanced computational music analysis, and grant broader accessibility to extensive musical archives~\cite{maria_alfaro_contreras_ismir}. Traditionally, OMR methods were predominantly rule-based, involving steps such as staff line removal and primitive detection algorithms, which posed significant practical difficulties \cite{rebelo2012optical}. However, recent advancements in Deep Learning (DL) have successfully addressed many of these long-standing obstacles, leading to substantial improvements in OMR performance \cite{shatri2020optical}. This is evidenced by the shift in the field towards end-to-end methods that attempt to solve the problem in a few steps, either by directly detecting musical objects in full images \cite{tuggener2024real}, or with single-step transcription pipelines at both the region \cite{icdar/MayerSHP24,rios2024sheet} and page levels \cite{fppiano/riosvila}.

Given these advances and extensive ongoing research, one might consider OMR a mature and well-developed subfield of MIR. Nevertheless, unlike other MIR tasks, OMR still lacks a comprehensive, high-quality benchmark corpus, complicating rigorous performance assessment and qualitative comparison between systems. Such benchmarks have significantly benefited other MIR areas; notable examples include MAPS~\cite{maps_dataset}, ASAP~\cite{ismir/FoscarinMRJS20}, and MAESTRO~\cite{maestro_dataset} in automatic music transcription; Ballroom~\cite{Ballroom_dataset} and GTZAN~\cite{gtzan_dataset} in beat tracking; or NSynth~\cite{nsynth_dataset} for instrument classification, to name a few. These established benchmarks have greatly facilitated standardized comparisons and the systematic evaluation of novel approaches.

To date, several attempts have been made to establish an OMR benchmark capable of addressing this gap. In their seminal work, Byrd and Simonsen proposed the OMRTestCorpus~\cite{byrdtestbed}; however, this corpus consists of only thirty-four pages with substantial graphical variability, rendering it insufficient and unsuitable for contemporary DL-based OMR approaches. Similarly, CVC-MUSCIMA~\cite{ijdar/FornesDGL12/cvcmuscima} is only devised for the staff-line removal step, while its extension MUSCIMA++~\cite{icdar/HajicP17b/muscima} is object detection-oriented, with annotations not representing full musical scores. The PrIMuS dataset, although introduced as the first end-to-end OMR dataset, includes only monophonic samples at the staff-region level~\cite{ismir/Calvo-ZaragozaR18}. For pianoform end-to-end recognition, the GrandStaff dataset~\cite{fppiano/riosvila} was presented, but it is synthetically generated and lacks sufficient diversity, as the musical excerpts primarily originate from a limited set of composers. Recently, the OLiMPiC dataset~\cite{icdar/MayerSHP24} has been proposed for pianoform recognition tasks. Nevertheless, it also exhibits limited variety regarding repertoire and musical purposes, a restricted size, and is not truly end-to-end, since it relies on Linearized XML, a compact representation of MusicXML which requires additional processing to obtain a fully renderable music score.

Considering the previously identified limitations and the existing gap in the field, in this work we introduce the \emph{Sheet Music Benchmark} (SMB), a novel, full-page end-to-end dataset specifically designed for benchmarking modern DL-based OMR systems. SMB supports comprehensive evaluation, including layout analysis, region-level recognition, and full-page transcription. In addition to the dataset, we design a new evaluation metric: the \emph{OMR Normalized Edit Distance} (OMR-NED). This metric pursues two main objectives: (i) summarizing the overall performance of an OMR system in a single numerical value, and (ii) enabling detailed profiling of OMR system errors by categorizing them into specific notation elements such as notes, rests, and measures, among others. To the best of our knowledge, the combination of SMB dataset and OMR-NED metric will significantly advance OMR research, similar to how established benchmarks have driven progress in other MIR areas, providing researchers and practitioners with a standardized and effective framework to rigorously evaluate OMR systems.

\section{Building the benchmark}\label{sec:building}

SMB corpora is built upon KernScores \cite{ismir/Sapp05}, an online library of musical data encoded in \texttt{**kern}. This encoding format allows further processing with the Humdrum Toolkit for Music Research.\footnote{\href{https://www.humdrum.org}{https://www.humdrum.org}} KernScores provides multiple representations for each piece, including a public-domain scan of the sheet music (when available), the \texttt{**kern} encoding, and a MIDI file. To build SMB, all the steps were performed by musically trained experts to ensure the quality of the annotations, such as the selection of the pieces, which included the \textit{linking} process between the sheet music scan and its corresponding \textit{raw} \texttt{**kern} encoding.\footnote{Here \textit{linking} means that when the scan was not available, annotators looked for it in different sources.}

\subsection{Annotation process}
\begin{figure} 
    \centering
    \includegraphics[alt={Labeled piece example. In blue the bounding boxes that locate the regions of interest. In the top-right corner the \textit{raw}\texttt{**kern} encoding of the first region of the page.},width=0.9\linewidth]{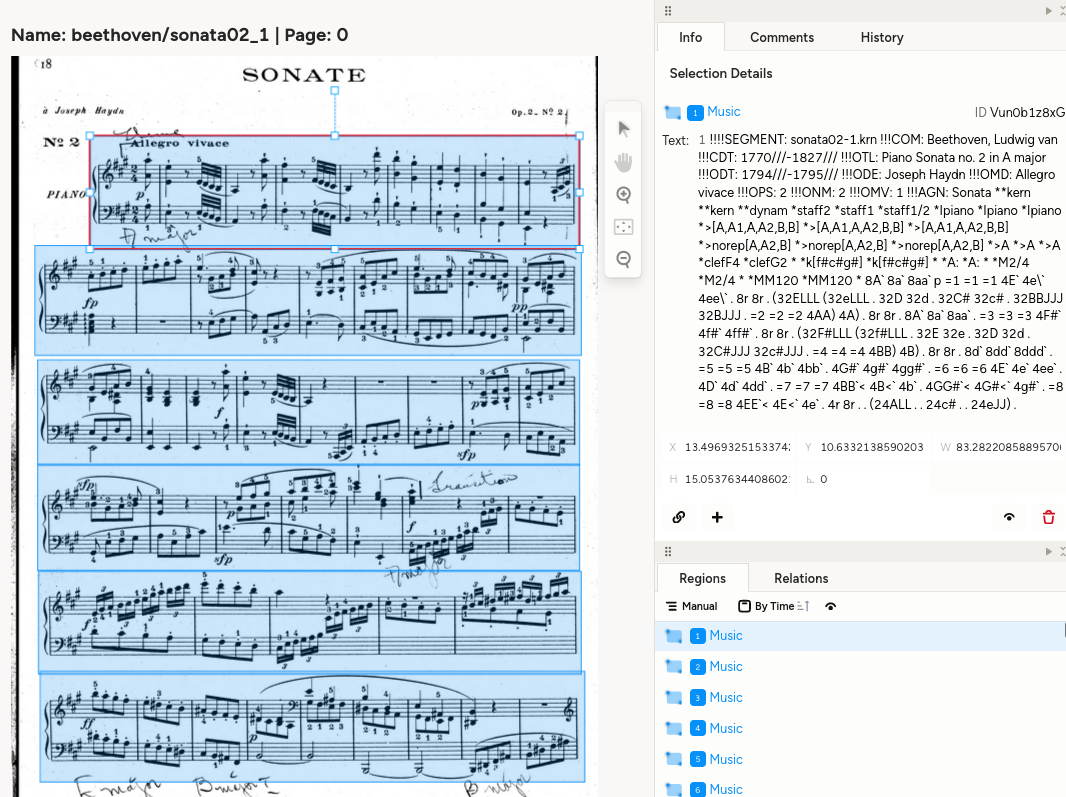}
    \caption{Labeled piece example. In blue the bounding boxes that locate the regions of interest. In the top-right corner the \textit{raw} \texttt{**kern} encoding of the first region of the page.}
    \label{fig:humansignal-labeling}
\end{figure}

The scanned scores were uploaded to HumanSignal,\footnote{\href{https://humansignal.com/}{https://humansignal.com/}} a web platform specifically designed for data annotation. Each musical  piece was manually labeled, starting with the assignment of a texture tag---Monophonic\footnote{Most of the samples are single-staff and single-voice however homophony and polyphony can appear.}, Pianoform, PianoAndVoice, Quartet, or Other.\footnote{Given the number of samples in the PianoAndVoice and Other texture tag, they will be referred to as Other for the rest of the paper.} Then, for each page, the regions corresponding to individual staves or musical systems were delimited by bounding boxes and labeled with their content in \textit{raw} \texttt{**kern} format (see Fig. \ref{fig:humansignal-labeling}).

These annotations were performed at the staff level (sometimes referred to as line-level annotations). While full-page end-to-end digitization is the ultimate goal of any OMR system, it remains a highly complex challenge for most architectures. Nonetheless, SMB also provides \textit{``diplomatic''}\footnote{Annotations indicate where the staff breaks are in the source score.} full-page annotations, suitable for staff-level and page-level transcription. These detailed annotations also allow \ac{LA} processes, making SMB appropriate for any modern OMR pipeline.

\subsection{The use of Humdrum \texttt{**kern}}
Music can be digitally represented in various formats, including Humdrum \texttt{**kern,} MEI, or MusicXML. However, \texttt{**kern} remains the most efficient and least verbose option, making it particularly suitable for \ac{OMR} DL-based systems. This efficiency, combined with the extensive Humdrum ecosystem—--which includes rendering tools like Verovio Humdrum Viewer~\cite{PuginZR14} and symbolic analysis frameworks such as humdrum-tools~\cite{humdrumr}--- as well as partitura~\cite{partitura_mec} and music21~\cite{CuthbertA10music21}, ensures \texttt{**kern} is both accessible and highly advantageous for \ac{OMR} applications.\footnote{The use of \texttt{**kern} for OMR does not constrain the later usage of another encoding format.}

\subsection{Postprocessing} \label{subsec:postprocessing}
\begin{figure}
    \centering
    \includegraphics[alt={Possible \texttt{**kern} ambiguous representations for an $A\#_{4}$ eighth note.},width=\linewidth]{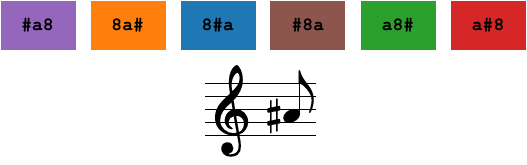}
    \caption{Possible \texttt{**kern} inconsistent representations for an $A\#_{4}$ eighth note.}
    \label{fig:kern_ambiguity}
\end{figure}

Once the annotation process was completed, compiling a robust, functional dataset for modern end-to-end \ac{OMR} methods posed significant challenges. An important component in the extracted \textit{raw} \texttt{**kern} fragments from KernScores is their fidelity to the original score, providing metadata such as the source, authorship, publication year, work titles, and more. However, for some DL experiments, the availability of this information together with the inherent ambiguity of the \textit{raw} \texttt{**kern} encoding itself---which can represent the same symbol in multiple ways (see Fig. \ref{fig:kern_ambiguity})---supposes a major drawback, and thus it needed to be addressed.

For that reason, the construction of standardized, high-quality \texttt{**kern} files, e.g., by constraining the representations of notes to reduce vocabulary size, is a key point of the SMB. The kernpy~\cite{kernpy_mec_2025} package was a major pillar during the postprocessing process providing features that converted non-standardized fragmented \texttt{**kern} data---as it comes from the labeling process---into complete standardized encoding ready to use by the Humdrum ecosystem tools: including the proper \texttt{**kern} headers, spine paths\footnote{\href{https://www.humdrum.org/Humdrum/representations/kern.html\#Spine\%20Paths}{https://www.humdrum.org/Humdrum/representations/kern.html}}and changes in clefs, key signatures, and time signatures. 

In addition, this package was also used to enforce structural consistency and perform semantic tokenization across different music categories. This approach even allows for the possibility of performing custom selections for specific research tasks, such as enabling the use of a tailored set of categories or avoiding the inclusion of articulations, accidentals, dynamics, or pitches. Consequently, three formats emerged from this step:


\begin{itemize}
    \item \textbf{\textit{raw}}. Represents the original fragmented \textit{raw} \texttt{**kern} KernScores encoding used for in the labeling steps without any change. 
    \item \textbf{\textit{kern}}. Standardized \texttt{**kern} file retrieved from the tokenized version obtained from kernpy package. Compatible with the Humdrum ecosystem toolkit.
    \item \textbf{\textit{ekern}}. The "Extended \texttt{**kern}" is a standardized tokenized version of \texttt{**kern} where each musical symbol is spaced by ``@'' and ``·'' dividing semantic token categories of each symbol, i.e., duration, pitch or accidental. This approach reduces significantly the vocabulary size of the dataset, although increasing sequence lengths.
\end{itemize}

Following automated data processing, the SMB underwent manual inspection at page and region levels to ensure error-free \texttt{**kern} and \texttt{**ekern} annotations. Encoding irregularities in problematic annotations were systematically solved.

\subsection{Data analysis}
In this subsection, we present a comprehensive analysis of SMB samples to provide insights into its features, distribution, and content. We depict key statistical properties, explore patterns within the data, and identify notable details given the standardized \texttt{**kern} encoding. This analysis serves to both validate the quality of the benchmark and illustrate in a clearer way the content of it.

SMB spans 4039 regions which correspond to 685 pages. Table \ref{tab:generalStatistics} provides a detailed description of the main features of this assortment in \texttt{**kern} format. As shown, the most representative texture is Pianoform. On one hand, it is one of the instruments with the broadest repertoire; on the other, it remains one of the most complicated textures to transcribe. Another point to highlight is that when increasing the number of instruments, i.e., Quartet texture, musicians do not usually play with the full score as they prefer to use parts. As a result, the number of annotated pages is lower than others. 

An interesting outcome is that the graphical features of higher-voice-count pages such as Quartet compared to Monophony, cause a drop in the number of regions by page, but the number of tokens by page remains similar. A key difference we find is the number of tokens per region as when the number of voices increases, the count rises drastically.

\begin{figure} 
    \centering
    \includegraphics[alt={Pitch token frequency distribution in SMB.},width=\linewidth]{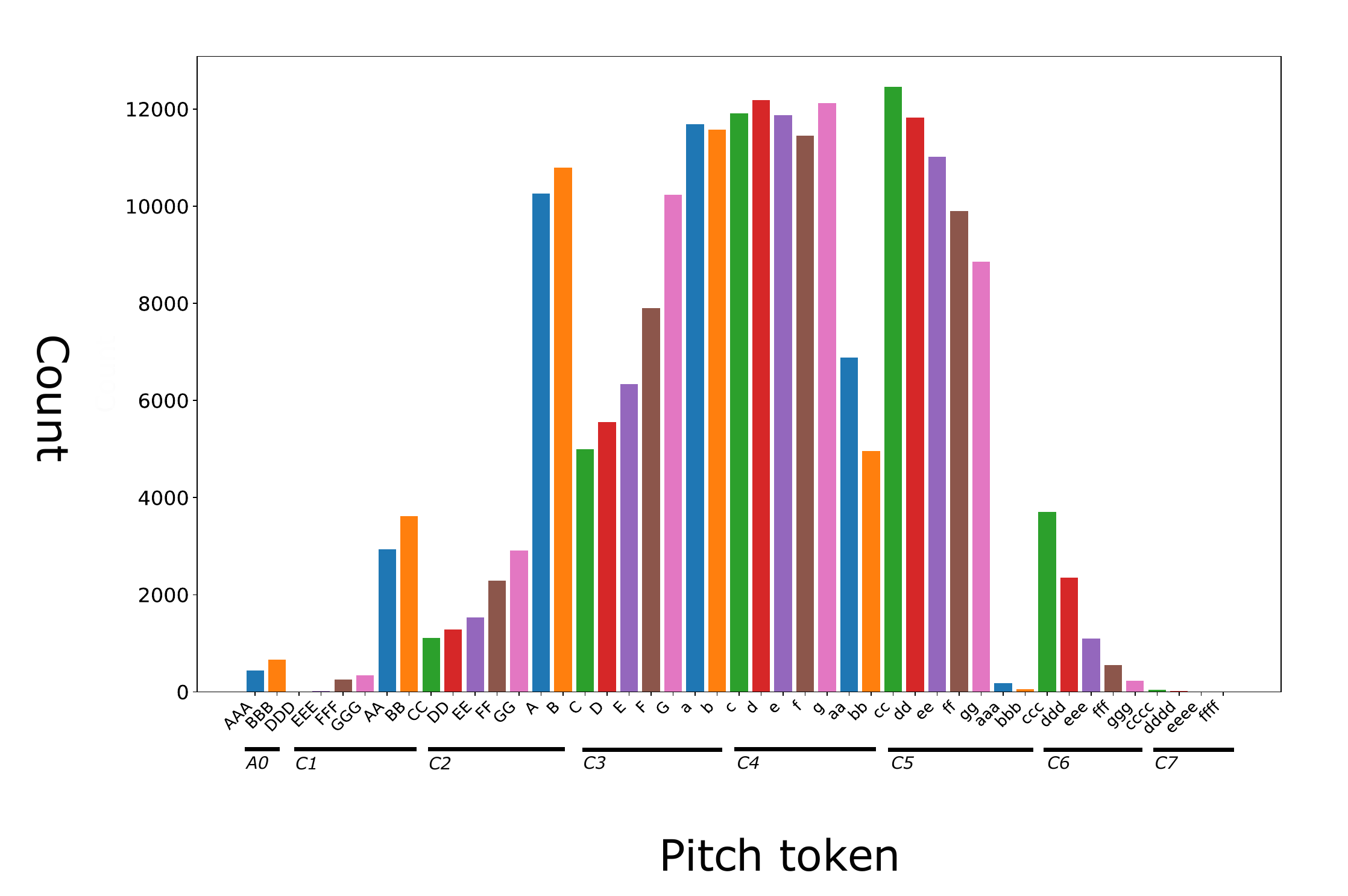}
    \caption{Pitch token frequency distribution in SMB.}
    \label{fig:pitch-historiogram.pdf}
\end{figure}

In Figure \ref{fig:pitch-historiogram.pdf} we show the Pitch token frequency distribution of SMB, where the most common pitches are the ones present between $A_{3}$ and $G_{5}$.

Different composers, styles, and periods were considered to increase the variability in the data. The heterogeneity of the collection spans from the Baroque period to the Ragtime eras. In between, there is a set of various musical periods such as the Classical, the Romantic, or the Impressionist, which provide a wide range of musical pieces. Some of the available composers are: Carl Maria von Weber, Domenico Scarlatti, Frédéric Chopin, Joseph Haydn, W. A. Mozart, L. van Beethoven, and Scott Joplin, among others.

The diversity present in SMB makes it to our best knowledge the first and most complete dataset in \ac{CWMN} for \ac{OMR} DL-based methodologies benchmarking, providing a set of diverse and complex scores.
 
\begin{table*}
    \centering
    \begin{tabular}{l c c c c c}
        \toprule
        Texture & Pages & Regions & Regions by page $(\mu \pm \sigma)$ & Tokens by page $(\mu \pm \sigma)$ & Tokens by region $(\mu \pm \sigma)$ \\
        \midrule
        Monophony  & 115 & 972 & \(8.4 \pm 3.5\) & \(1308.5 \pm 719.7\) & \(165.2 \pm 51.1\) \\        
        Pianoform   & 469 & 2686 & \(5.7 \pm 1.1\) & \(1371.3 \pm 679.4\) & \(263.5 \pm 105.0\) \\
        Quartet  & 79 & 304 & \(3.8 \pm 0.5\) & \(1338.4 \pm 413.2\) & \(373.1 \pm 92.9\) \\
        Other & 22 & 77 & \(3.5 \pm 1.5\) & \(1203.0 \pm 671.8\) & \(349.8 \pm 136.4\) \\
        \midrule
        All & 685 & 4039 & \(5.8 \pm 2.2\) & \(1351.6 \pm 660.9\) & \(249.7 \pm 110.2\) \\
        \bottomrule
    \end{tabular}
    \caption{Results of SMB page, region and token count analysis in \texttt{**kern} encoding.}
    \label{tab:generalStatistics}
\end{table*}

\subsection{Publication}
Ensuring accessibility and ease of use for the research community, SMB is available at Hugging Face Datasets platform.\footnote{\href{https://huggingface.co/datasets/PRAIG/SMB}{huggingface.co/datasets/PRAIG/SMB}}

\section{Evaluation metric} \label{sec:evaluation_metric}
One key aspect of building a successful OMR system is the way of measuring the quality of it. Measuring OMR performance remains an open problem due to a set of unanswered questions: \textit{What is considered a music symbol? How do we measure the correction effort?} The answer to these questions remains an open topic that is being addressed in conjunction with musicologists.

Until today, the most popular metric used when evaluating OMR has been the \textit{Symbol Error Rate} (i.e. in some works referred to as Edit Distance, Character Error Rate, or Music Error Rate), that is computed as the average number of elementary operations (insertions, deletions, or substitutions) required to convert prediction $\hat{\mathbf{z}}_{i}$ into reference $\mathbf{z}_{i}$, normalized by the length of the latter.

Indeed, it still stands as the best metric to correlate the human correction effort that is necessary to obtain the desired score. However such correction effort does not indicate in which aspect of the transcription process the OMR is struggling, e.g., clefs, note heads, beams, accidentals, key signatures, or anything else.

In this work, we leverage Music-Score-Diff \cite{foscarin2019diff} currently maintained by Greg Chapman and better known as MusicDiff~\footnote{\href{https://github.com/gregchapman-dev/musicdiff}{github.com/gregchapman-dev/musicdiff}}. MusicDiff computes the visual notation differences between two music scores; however, this tool has been known for having trouble processing the scores from OMR pipelines because it requires a parseable score (which is certainly not always the case with OMR systems). Due to this, and other shortcomings, we have improved MusicDiff significantly so it can now do the following: (i) it can compare non-parseable predicted \texttt{**kern} scores, (ii) it can compare all musical objects, not just notes/rests, (iii) it can compare all the notes in a measure without regard to voicing and/or membership in chords, and (iv) it can be told at what fine-grained level of detail it should compare the scores.  We have also significantly changed how MusicDiff computes edit distance, and we have carefully audited and changed how music symbols are defined/counted.

Hence we present the resultant metric score, referred to as the OMR Normalized Edit Distance (OMR-NED), designed specifically to allow visual score comparison and measuring in a fine-grained manner, therefore standing as the first alternative to the traditionally used SER. OMR-NED defines a set of categories, and a detailed list of visual music symbols for each category, which can then be compared.  Note that these music symbols are not defined in terms of \texttt{**kern} (or \texttt{**ekern}). Instead, they are defined in terms of visual music notation, in a file format agnostic way. The OMR-NED metric for the score comparison can be computed as the average number of \textit{music symbol} insertions and deletions required to convert the predicted score into the reference score, normalized by the sum of the number of \textit{music symbols} in both:

\begin{equation}
\label{eq:omrned}
    \text{OMR-NED}=\frac{I+D}{N_1+N_2}
\end{equation}

$I$ and $D$ are the total number of insertions and deletions of individual music symbols for all categories, and $N_1$ + $N_2$ is the total number of music symbols in the predicted and ground truth scores.  We normalize using both scores' music symbol counts, because when a symbol is changed (say there is a predicted \dynmark{2/4} time signature, but the ground truth is a \dynmark{3/4} time signature), instead of computing a substitution of one symbol, we compute a deletion of the ``2'' and an insertion of the ``3'', for an edit distance of two symbols.

MusicDiff can be pointed at a folder full of ground truth (reference) score files, and a folder full of same-named predicted score files.  When running in this mode, MusicDiff will produce a CSV file containing a spreadsheet with a row for each score file comparison and a column for each category.  Each column contains the edit distances for that category, with the total edit distance and OMR-NED metrics for all the score comparisons in the final columns.  There is a summary row for the entire run of score comparisons at the bottom, with a total edit distance for each category across all the scores, and an overall OMR-NED metric for the entire run.

\subsection{OMR-NED categories}
In this subsection we present the categories related to music notes and rests and the non-notes ones. This will be further emphasized in the fine-grained analysis that OMR-NED depicts.

\textbf{Notes/Rests:}\footnote{The description here is of the default categories. There can be more symbols per category (and more categories) if Style and/or Metadata is requested in addition to the default detail level. See \href{https://gregchapman-dev.github.io/musicdiff/musicdiff/detaillevel.html}{gregchapman-dev.github.io/musicdiff/musicdiff/detaillevel.html}
 for more information about specifying detail levels to MusicDiff.} Notes and rests contain multiple categories, each with its own symbol count.  MusicDiff only directly compares notes/rests that have the same pitch and are on the same exact offset within the measure.  This means that notes that are mismatched by pitch (or by offset within the measure) will have a larger edit distance because the entire note (and all its symbols) is deleted, and then the new note is inserted.

\textbf{\textit{Available categories:}} \textbf{Pitch} — One symbol for the pitch. \textbf{Accidental} — One symbol for the accidental. \textbf{Tie} — One symbol if the note is tied to a subsequent note. \textbf{Note head} — One symbol for the note head type (quarter note, half note, whole note, etc).  The note head type for eighth notes and smaller is still a quarter note head.  The flags/beams category differentiates further between shorter notes. \textbf{Note flags/beams} — One symbol per flag or beam. \textbf{Dots} — One symbol per dot. \textbf{Articulations} — One symbol per articulation (staccato, tenuto, etc). \textbf{Ornaments} — One symbol per ornament (trill, mordent, etc). \textbf{Grace} — One symbol if a grace note; one more symbol if slashed.

\textbf{Non-Notes:} Non-note objects are a single category each.  MusicDiff only directly compares non-note objects that are at the same exact offset within the measure.  So, for example, text directions that are shifted horizontally from each other are deleted/inserted, instead of returning the edit distance between the strings.

\textbf{\textit{Available categories:}} \textbf{Dynamic} — Dynamics such as \dynmark{p}, \dynmark{fff}, etc are one symbol only (not one symbol per character).  Hairpin dynamic markings are one symbol for the hairpin direction plus an extra symbol for the hairpin duration. \textbf{Clef} — Clefs are one symbol. \textbf{Key signature} — Key signatures are one symbol per accidental. \textbf{Time signature} — Time signatures are one symbol for the top, one for the bottom (e.g. \dynmark{12/8} is two symbols, \dynmark{C} is one symbol). \textbf{Slur} — One symbol for duration. \textbf{Ottava} — Two symbols: one for ottava type (e.g. \dynmark{8va}, \dynmark{8ba}, etc), and one for duration. \textbf{Direction} — One symbol for each character in the string. \textbf{Arpeggio} — One symbol for arpeggio type (up, down, undirected, non-arpeggiated). One more symbol if arpeggio spans multiple staves. \textbf{Chord symbol} — One symbol. \textbf{Lyric} — One symbol for each character in the lyric syllable. One symbol for verse number/identifier. \textbf{Ending} — One symbol for each character in the ending name. One symbol for measure count.

\subsection{Edit distances}
Edit lists are calculated as a series of inserts/deletes of musical objects, with the edit distance being the total number of symbols inserted/deleted. For example, if a predicted note has a flag, but the matching ground truth note has a beam instead, the edit distance will be 2: delete the flag (one symbol) and insert the beam (one symbol).

At a slightly higher level, if the predicted score has a note where the ground truth score has a rest, the edit will delete the note, and insert the rest.  The edit distance in that case will be the number of symbols in the deleted note plus the number of symbols in the inserted rest.

At an even higher level, if the predicted score has an extra measure that does not exist in the ground truth score, that measure will be deleted, with an edit distance equal to the total number of symbols in that predicted measure.

\section{Baseline results} \label{sec:baseline}
This work constitutes the first to introduce both the SMB benchmark dataset and the fine-grained metric evaluation OMR-NED. Therefore, in this section, we report a performance baseline using a state-of-the-art model, which serves as a reference for future work.

While the recommended use of SMB is as test set, for the baseline we consider each texture type a subset of samples, namely Monophony, Pianoform, Quartet, and Other. These scores are used at the region level (assuming a previous \ac{LA} step) and following a 5-fold cross validation framework, ensuring that every region takes part in the test split once. We train the framework both with \texttt{**kern} and \texttt{**ekern} encoding.

For the learning framework, we resort to the state-of-the-art Sheet Music Transformer \cite{rios2024sheet}. This architecture consists of a \textit{encoder-decoder} network. The \emph{encoder} is in charge of retrieving the image features and the \emph{decoder} is a conditioned language model that predicts the most probable music symbol in an auto-regressive fashion.

The model was trained without the use of any pretrained weights. We iterate for 400 epochs, considering the ADAM optimizer with a fixed learning rate of $10^{-4}$. We keep the weights that minimize the SER metric in the validation partition, validating every 5 epochs. Finally, all experiments were run using the Python language (v. 3.12.3) with the PyTorch framework (v. 2.0.0) on a single NVIDIA RTX 4080 card with 20GB of video memory.

Table \ref{tab:results3} presents the average results obtained using the considered experimental baseline in terms of SER and the introduced OMR-NED. Regarding the (low) recognition rates, it is worth noting that Transformer architectures typically require large corpora to achieve convergence. Nevertheless, we report these values to facilitate future research using SMB. The results indicate ample room for improvement, highlighting the dataset as a challenging benchmark that should drive advancements in OMR.

\begin{table}[!ht]
    \centering
    \resizebox{0.9\linewidth}{!}{
    \begin{tabular}{llcc}
        \toprule[1pt]
        \multirow{1}{*}{Texture} & \multirow{1}{*}{Encoding} & \multicolumn{1}{c}{ $\downarrow\,$OMR-NED (\%)} & \multirow{1}{*}{\makecell{ $\downarrow\,$SER (\%)}} \\
        
        \cmidrule(lr){1-4}
        \multirow{2}{*}{Monophony} 
                                    & \texttt{kern}    & $94.1 \pm 5.0$ & $57.1 \pm 0.8$\\
                                    & \texttt{ekern}     & $98.8 \pm 1.0$ & $65.8 \pm 1.2$\\

        \cmidrule(lr){1-4}
        \multirow{2}{*}{Pianoform}  & \texttt{kern} & $57.4 \pm 6.0$ & $31.4 \pm 1.5$\\
                                    & \texttt{ekern} & $77.2 \pm 2.0$ & $55.1 \pm 2.2$\\        
        \cmidrule(lr){1-4}
        \multirow{2}{*}{Quartet}  & \texttt{kern}  & $92.3 \pm 1.2$ & $39.8 \pm 1.6$\\
                                    & \texttt{ekern}   & $93.3 \pm 1.1$ & $82.9 \pm 1.2$\\
        \cmidrule(lr){1-4}
        \multirow{2}{*}{Other}  & \texttt{kern} & $89.6 \pm 2.4$ & $60.2 \pm 6.2$\\
                                    & \texttt{ekern}  & $93.7 \pm 2.0$ & $72.4 \pm 7.3$ \\
        \bottomrule[1pt]
    \end{tabular}
    }
\caption{Results in terms of the SER (\%) and OMR-NED (\%) metrics when considering Monophony, Pianoform, Quartet and Other score textures. All cases are evaluated using 5-fold cross validation.}
\label{tab:results3}
\end{table}

\section{Case study: OMR-NED \textit{vs.} SER} \label{sec:case_study}

\begin{figure}
    \centering
    \includegraphics[alt={Monophonic (with lyrics) region-level transcription example. Red boxes and circles represent transcription errors.},width=0.85\linewidth]{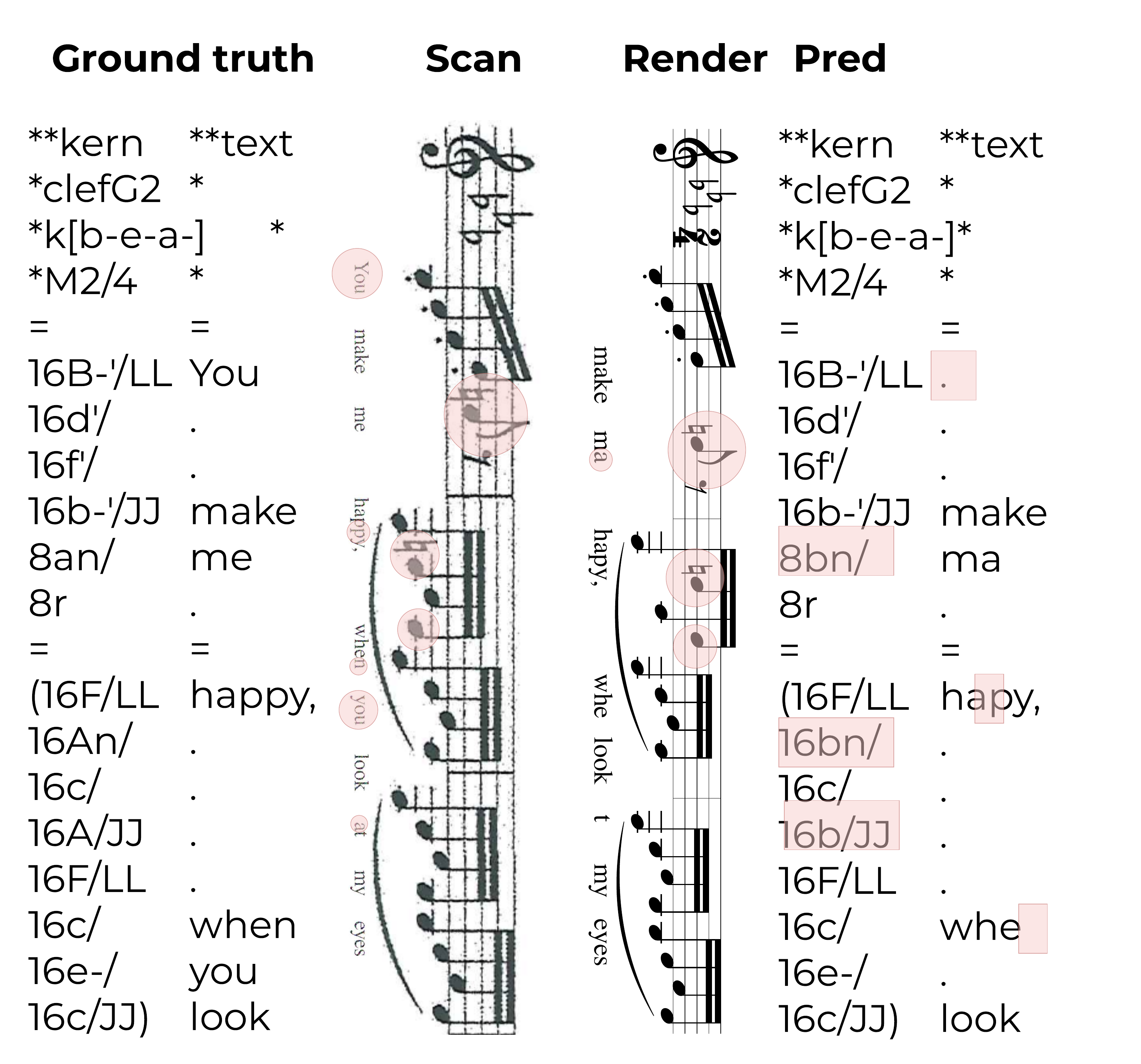}
    \caption{Monophonic (with lyrics) region-level transcription example. Red boxes and circles represent transcription errors.}
    \label{fig:example_1_case_study}
\end{figure}

To better exhibit the pros and cons of using OMR-NED (at the region level) compared to traditionally employed SER, we study two different transcription examples. The first one (see Fig. \ref{fig:example_1_case_study}) highlights the wrongly transcribed notes and lyrics in a monophonic score. If we look at the results obtained in Table \ref{tab:case_study} the different OMR-NED Categories permit the fine-grained analysis of this behavior. With 65\% of the errors related to notes and 35\% to lyrics mistranscriptions.

\begin{figure}
    \centering
    \includegraphics[alt={Pianoform region-level transcription example. Red boxes and circles represent transcription errors.},width=\linewidth]{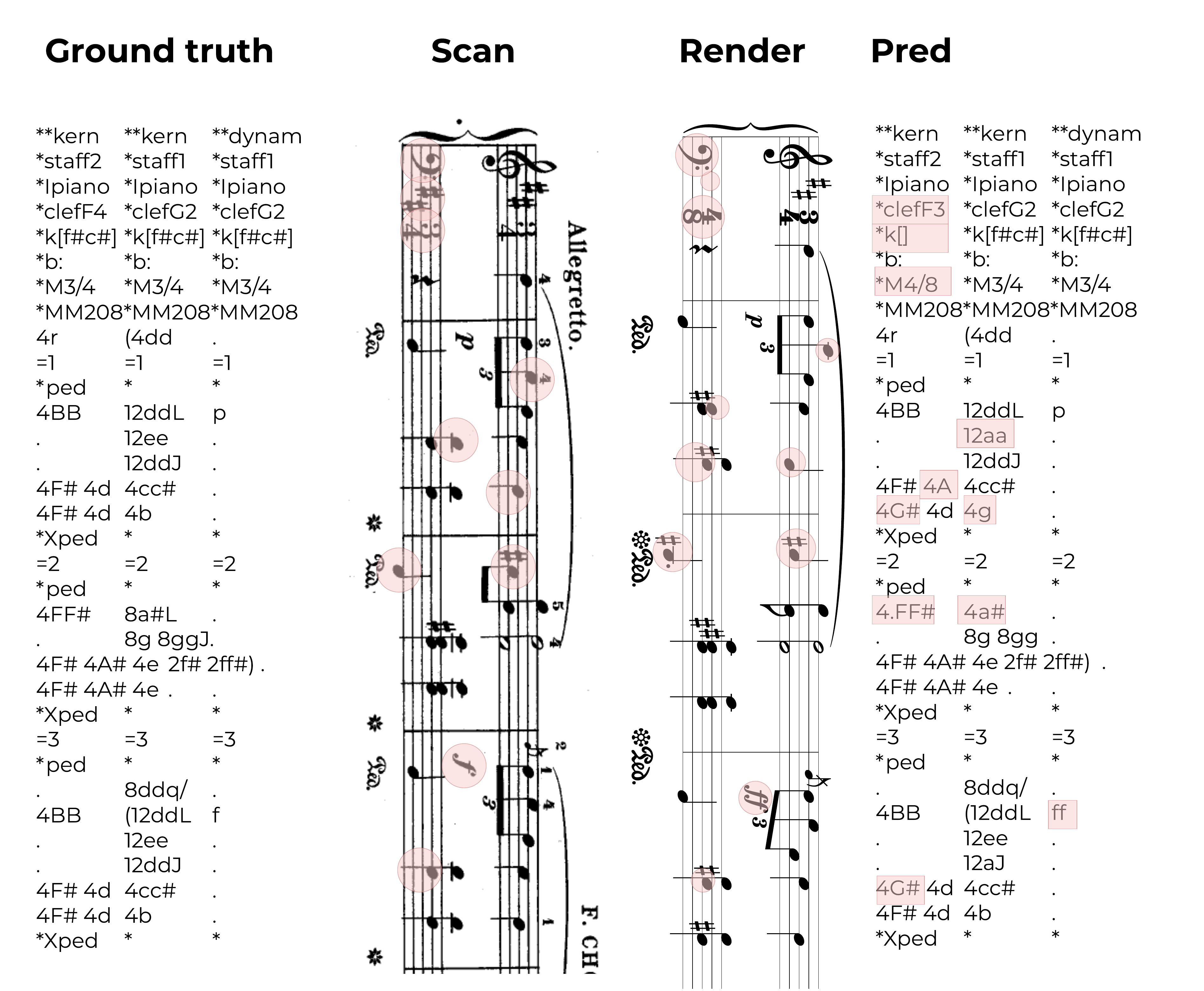}
    \caption{Pianoform region-level transcription example. Red boxes and circles represent transcription errors.}
    \label{fig:example_2_case_study}
\end{figure}

Figure \ref{fig:example_2_case_study} shows an example of a pianoform transcription. In this case, most of the errors (88.2\%) correspond to the note category, however, OMR-NED outlines problems in the extra category, indicating that there are also mistaken music symbols that comprehend dynamics or left-hand clef and key. It is worth mentioning that given the level of detail OMR-NED depicts and the procedure to calculate the different edit distances between hypothesis and ground truth (see Section \ref{sec:evaluation_metric}), the overall score penalizes the transcription errors more than traditionally used SER. 

The comparative analysis of OMR-NED and SER through these case studies demonstrates the advantages of adopting a more refined and structured evaluation metric. While SER provides an acceptable measurement of transcription accuracy, OMR-NED enables a more granular understanding of the specific transcription error types. This is particularly evident in the first example, where OMR-NED differentiates between note and lyric errors. Similarly, the second example highlights OMR-NED's ability to capture complex transcription issues in polyphonic pianoform music, particularly by exposing errors in musical symbols beyond pitch content—such as dynamics and clef misclassifications. OMR-NED as a more informative metric, encompasses the development of more robust OMR systems.

\begin{table}[!ht]
    
    \centering
    \resizebox{\linewidth}{!}{
    \begin{tabular}{lcccccccc}
        \toprule[1pt]
        \multirow{2}{*}{Example}  & \multicolumn{7}{c}{OMR-NED Categories} & \multirow{2}{*}{\makecell{Symbol\\Error Rate}} \\
        \cmidrule(lr){2-8}

         & Note & Extra & Lyrics & Measure & Part & StaffGroup & OMR-NED & \\
        \cmidrule(lr){1-7} \cmidrule(lr){8-8} \cmidrule(lr){9-9}

        Fig. \ref{fig:example_1_case_study} &  65.0   &    0    &   35.0 &     0 &     0      &      0  & 13.6 & 9.0\\

        Fig. \ref{fig:example_2_case_study} &  88.2   &   11.8     &   0 &     0 &     0      &      0  & 40.9 & 13.0 \\

        \bottomrule[1pt]
    \end{tabular}
    }
\caption{Results in terms of the SER (\%) and OMR-NED Categories (\%) metrics. Note that OMR-NED is prepared for full-page recognition thus some categories do not affect the OMR-NED score in the region-level scenario.}
\label{tab:case_study}
\end{table}

\section{Conclusion} \label{sec:conclusion}
In this work, we addressed a long-standing gap in the Optical Music Recognition (OMR) literature by introducing the \emph{Sheet Music Benchmark (SMB)}, a collection of scores specifically designed for evaluating modern end-to-end OMR pipelines. SMB constitutes the first publicly available corpus that allows comprehensive end-to-end evaluation at multiple levels, including layout analysis as well as region-level and full-page transcription tasks.  We also provided baseline results to guide future research efforts in this area. The reported performance with a state-of-the-art methodology demonstrate the benchmark’s complexity and its effectiveness in assessing OMR systems targeting Common Western Modern Notation. 

Furthermore, we introduced a novel evaluation metric, the \emph{OMR Normalized Edit Distance} (OMR-NED), which enables a fine-grained analysis of OMR system performance. OMR-NED decomposes the traditional edit distance calculation into distinct error categories, facilitating precise profiling of potential system deficiencies. As demonstrated in Sec.~\ref{sec:case_study}, the use of OMR-NED provides valuable insights into transcription errors, thereby enabling targeted improvements in OMR frameworks.

The combined contributions of SMB and OMR-NED establish a standardized evaluation pipeline, facilitating clear performance comparisons among OMR systems. Thus, this work significantly advances the field by addressing a critical need within the OMR research community.


\section{Acknowledgments}
This paper is supported by grant CISEJI/2023/9 from ``Programa para el apoyo a personas investigadoras con talento (Plan GenT) de la Generalitat Valenciana''.

\bibliography{bibliography}

@inproceedings{foscarin2019diff,
    title={A diff procedure for music score files},
    author={Foscarin, Francesco and Jacquemard, Florent and Fournier-S’niehotta, Raphael},
    booktitle={6th International Conference on Digital Libraries for Musicology},
    pages={58--64},
    year={2019}
}

@article{Ballroom_dataset,
  author       = {Fabien Gouyon and
                  Anssi Klapuri and
                  Simon Dixon and
                  M. Alonso and
                  George Tzanetakis and
                  C. Uhle and
                  Pedro Cano},
  title        = {An experimental comparison of audio tempo induction algorithms},
  journal      = {{IEEE} Trans. Speech Audio Process.},
  volume       = {14},
  number       = {5},
  pages        = {1832--1844},
  year         = {2006},
}

@inproceedings{maestro_dataset,
  author       = {Curtis Hawthorne and
                  Andriy Stasyuk and
                  Adam Roberts and
                  Ian Simon and
                  Cheng{-}Zhi Anna Huang and
                  Sander Dieleman and
                  Erich Elsen and
                  Jesse H. Engel and
                  Douglas Eck},
  title        = {Enabling Factorized Piano Music Modeling and Generation with the {MAESTRO}
                  Dataset},
  booktitle    = {7th International Conference on Learning Representations, {ICLR} 2019,
                  New Orleans, LA, USA, May 6-9, 2019},
  publisher    = {OpenReview.net},
  year         = {2019},

}

@inproceedings{nsynth_dataset,
  author       = {Jesse H. Engel and
                  Cinjon Resnick and
                  Adam Roberts and
                  Sander Dieleman and
                  Mohammad Norouzi and
                  Douglas Eck and
                  Karen Simonyan},
  editor       = {Doina Precup and
                  Yee Whye Teh},
  title        = {Neural Audio Synthesis of Musical Notes with WaveNet Autoencoders},
  booktitle    = {Proceedings of the 34th International Conference on Machine Learning,
                  {ICML} 2017, Sydney, NSW, Australia, 6-11 August 2017},
  series       = {Proceedings of Machine Learning Research},
  volume       = {70},
  pages        = {1068--1077},
  publisher    = {{PMLR}},
  year         = {2017},
}

@article{gtzan_dataset,
  author       = {George Tzanetakis and
                  Perry R. Cook},
  title        = {Musical genre classification of audio signals},
  journal      = {{IEEE} Trans. Speech Audio Process.},
  volume       = {10},
  number       = {5},
  pages        = {293--302},
  year         = {2002},
}

@techreport{maps_dataset,
  TITLE = {{MAPS - A piano database for multipitch estimation and automatic transcription of music}},
  AUTHOR = {Emiya, Valentin and Bertin, Nancy and David, Bertrand and Badeau, Roland},
  TYPE = {Research Report},
  PAGES = {11},
    INSTITUTION = {-},
  YEAR = {2010},
  MONTH = Jul,
  HAL_ID = {inria-00544155},
  HAL_VERSION = {v1},
}

@incollection{Huron1997,
    address = {Cambridge, MA, USA},
    author = {Huron, David},
    booktitle = {Beyond MIDI: The handbook of musical codes},
    isbn = {0262193949},
    month = {jan},
    pages = {375--401},
    publisher = {MIT Press},
    title = {{Humdrum and Kern: Selective Feature Encoding BT  - Beyond MIDI: The handbook of musical codes}},
    year = {1997}
}

@inproceedings{ismir/FoscarinMRJS20,
  author       = {Francesco Foscarin and
                  Andrew McLeod and
                  Philippe Rigaux and
                  Florent Jacquemard and
                  Masahiko Sakai},
  editor       = {Julie Cumming and
                  Jin Ha Lee and
                  Brian McFee and
                  Markus Schedl and
                  Johanna Devaney and
                  Cory McKay and
                  Eva Zangerle and
                  Timothy de Reuse},
  title        = {{ASAP:} a dataset of aligned scores and performances for piano transcription},
  booktitle    = {Proceedings of the 21th International Society for Music Information
                  Retrieval Conference, {ISMIR} 2020, Montreal, Canada, October 11-16,
                  2020},
  pages        = {534--541},
  year         = {2020},
}

@article{byrdtestbed,
author = {Byrd, Donald and Simonsen, Jakob},
year = {2015},
month = {07},
pages = {},
title = {{Towards a Standard Testbed for Optical Music Recognition: Definitions, Metrics, and Page Images}},
volume = {44},
journal = {Journal of New Music Research},
}

@inproceedings{maria_alfaro_contreras_ismir,
  author       = {María Alfaro-Contreras and
                  David Rizo and
                  Jose M. Iñesta and
                  Jorge Calvo-Zaragoza},
  title        = {{OMR-assisted transcription: a case study with 
                   early prints}},
  booktitle    = {{Proceedings of the 22nd International Society for 
                   Music Information Retrieval Conference}},
  year         = 2021,
  pages        = {35-41},
  publisher    = {ISMIR},
  address      = {Online},
  month        = nov,
  venue        = {Online},
}

@article{tuggener2024real,
  title={Real world music object recognition},
  author={Tuggener, Lukas and Emberger, Raphael and Ghosh, Adhiraj and Sager, Pascal and Satyawan, Yvan Putra and Montoya, Javier and Goldschagg, Simon and Seibold, Florian and Gut, Urs and Ackermann, Philipp and others},
  journal={Transactions of the International Society for Music Information Retrieval},
  volume={7},
  number={1},
  pages={1--14},
  year={2024},
  publisher={Ubiquity Press}
}

@inproceedings{HankinsonRF11,
  author    = {Andrew Hankinson and
               Perry Roland and
               Ichiro Fujinaga},
  title     = {{The Music Encoding Initiative as a Document-Encoding Framework}},
  booktitle = {Proceedings of the 12th International Society for Music Information
               Retrieval Conference, {ISMIR} 2011, Miami, Florida, USA, October 24-28,
               2011},
  pages     = {293--298},
  publisher = {University of Miami},
  year      = {2011}
}

@inproceedings{icdar/HajicP17b/muscima,
  author       = {Jan Hajic and
                  Pavel Pecina},
  title        = {{The MUSCIMA++ Dataset for Handwritten Optical Music Recognition}},
  booktitle    = {14th {IAPR} International Conference on Document Analysis and Recognition,
                  {ICDAR} 2017, Kyoto, Japan, November 9-15, 2017},
  pages        = {39--46},
  publisher    = {{IEEE}},
  year         = {2017},
}

@article{Calvo-ZaragozaH20understanding,
  author       = {Jorge Calvo{-}Zaragoza and
                  Jan Hajic Jr. and
                  Alexander Pacha},
  title        = {{Understanding Optical Music Recognition}},
  journal      = {{ACM} Comput. Surv.},
  volume       = {53},
  number       = {4},
  pages        = {77:1--77:35},
  year         = {2021},
}

@inproceedings{ismir/Sapp05,
  author       = {Craig Stuart Sapp},
  title        = {Online Database of Scores in the Humdrum File Format},
  booktitle    = {{ISMIR} 2005, 6th International Conference on Music Information Retrieval,
                  London, UK, 11-15 September 2005, Proceedings},
  pages        = {664--665},
  year         = {2005},
}

@inproceedings{partitura_mec,
  title={{Partitura: A Python Package for Symbolic Music Processing}},
  author={Cancino-Chac\'{o}n, Carlos Eduardo and Peter, Silvan David and Karystinaios, Emmanouil and Foscarin, Francesco and Grachten, Maarten and Widmer, Gerhard},
  booktitle={{Proceedings of the Music Encoding Conference (MEC2022)}},
  address={Halifax, Canada},
  year={2022}
}

@InProceedings{humdrumr,
  title = {humdrumR: a New Take on an Old Approach to Computational Musicology},
  author = {Nathaniel Condit-Schultz and Claire Arthur},
  booktitle = {Proceedings of the International Society for Music Information Retrieval},
  year = {2019},
  pages = {715--722},
}

@inproceedings{kernpy_mec_2025,
  title={{kernpy: a Humdrum **Kern Oriented Python Package for Optical Music Recognition Tasks}},
  author={Cerveto-Serrano, Joan and Rizo, David and Calvo-Zaragoza, Jorge},
  booktitle={{Proceedings of the Music Encoding Conference (MEC2025)}},
  address={London, United Kingdom},
  year={2025}
}

@inproceedings{icdar/MayerSHP24,
  author       = {Jir{\'{\i}} Mayer and
                  Milan Straka and
                  Jan Hajic and
                  Pavel Pecina},
  editor       = {Elisa H. Barney Smith and
                  Marcus Liwicki and
                  Liangrui Peng},
  title        = {{Practical End-to-End Optical Music Recognition for Pianoform Music}},
  booktitle    = {Document Analysis and Recognition - {ICDAR} 2024 - 18th International
                  Conference, Athens, Greece, August 30 - September 4, 2024, Proceedings,
                  Part {VI}},
  series       = {Lecture Notes in Computer Science},
  volume       = {14809},
  pages        = {55--73},
  publisher    = {Springer},
  year         = {2024},
}

@article{fppiano/riosvila,
  author       = {Antonio R{\'{\i}}os{-}Vila and
                  Jorge Calvo{-}Zaragoza and
                  David Rizo and
                  Thierry Paquet},
  title        = {{Sheet Music Transformer ++: End-to-End Full-Page Optical Music Recognition
                  for Pianoform Sheet Music}},
  journal      = {CoRR},
  volume       = {abs/2405.12105},
  year         = {2024},
  eprinttype    = {arXiv},
  eprint       = {2405.12105},
}

@inproceedings{rios2024sheet,
  title={Sheet music transformer: End-to-end optical music recognition beyond monophonic transcription},
  author={R{\'\i}os-Vila, Antonio and Calvo-Zaragoza, Jorge and Paquet, Thierry},
  booktitle={International Conference on Document Analysis and Recognition},
  pages={20--37},
  year={2024},
  organization={Springer}
}

@inproceedings{ismir/Calvo-ZaragozaR18,
  author       = {Jorge Calvo{-}Zaragoza and
                  David Rizo},
  editor       = {Emilia G{\'{o}}mez and
                  Xiao Hu and
                  Eric Humphrey and
                  Emmanouil Benetos},
  title        = {{Camera-PrIMuS: Neural End-to-End Optical Music Recognition on Realistic
                  Monophonic Scores}},
  booktitle    = {Proceedings of the 19th International Society for Music Information
                  Retrieval Conference, {ISMIR} 2018, Paris, France, September 23-27,
                  2018},
  pages        = {248--255},
  year         = {2018},
}

@inproceedings{CuthbertA10music21,
  author       = {Michael Scott Cuthbert and
                  Christopher Ariza},
  editor       = {J. Stephen Downie and
                  Remco C. Veltkamp},
  title        = {Music21: {A} Toolkit for Computer-Aided Musicology and Symbolic Music
                  Data},
  booktitle    = {Proceedings of the 11th International Society for Music Information
                  Retrieval Conference, {ISMIR} 2010, Utrecht, Netherlands, August 9-13,
                  2010},
  pages        = {637--642},
  publisher    = {International Society for Music Information Retrieval},
  year         = {2010},
}

@inproceedings{PuginZR14,
  author       = {Laurent Pugin and
                  Rodolfo Zitellini and
                  Perry Roland},
  editor       = {Hsin{-}Min Wang and
                  Yi{-}Hsuan Yang and
                  Jin Ha Lee},
  title        = {Verovio: {A} library for Engraving {MEI} Music Notation into {SVG}},
  booktitle    = {Proceedings of the 15th International Society for Music Information
                  Retrieval Conference, {ISMIR} 2014, Taipei, Taiwan, October 27-31,
                  2014},
  pages        = {107--112},
  year         = {2014},
}

@article{rebelo2012optical,
  title={Optical music recognition: state-of-the-art and open issues},
  author={Rebelo, Ana and Fujinaga, Ichiro and Paszkiewicz, Filipe and Marcal, Andre RS and Guedes, Carlos and Cardoso, Jaime S},
  journal={International Journal of Multimedia Information Retrieval},
  volume={1},
  pages={173--190},
  year={2012},
  publisher={Springer}
}

@inproceedings{good2001musicxml,
  title={{MusicXML: An internet-friendly format for sheet music}},
  author={Good, Michael and others},
  booktitle={Xml conference and expo},
  pages={03--04},
  year={2001},
  organization={Citeseer}
}

@article{ijdar/FornesDGL12/cvcmuscima,
  author       = {Alicia Forn{\'{e}}s and
                  Anjan Dutta and
                  Albert Gordo and
                  Josep Llad{\'{o}}s},
  title        = {{CVC-MUSCIMA:} a ground truth of handwritten music score images for
                  writer identification and staff removal},
  journal      = {Int. J. Document Anal. Recognit.},
  volume       = {15},
  number       = {3},
  pages        = {243--251},
  year         = {2012},
}

@article{shatri2020optical,
  title={Optical music recognition: State of the art and major challenges},
  author={Shatri, Elona and Fazekas, Gy{\"o}rgy},
  journal={arXiv preprint arXiv:2006.07885},
  year={2020}
}

\end{document}